# Generalized Back-Stepping Experience Replay in Sparse-Reward Environments

Guwen Lyu[1], and Masahiro Sato[1]

[1]Department of International Environment and Resources Policy, Graduate School of International Cultural Studies, Tohoku University, Sendai, Miyagi 980-8576, Japan

**ABSTRACT** Back-stepping experience replay (BER) is a reinforcement learning technique that can accelerate learning efficiency in reversible environments. BER trains an agent with generated back-stepping transitions of collected experiences and normal forward transitions. However, the original algorithm is designed for a dense-reward environment that does not require complex exploration, limiting the BER technique to demonstrate its full potential. Herein, we propose an enhanced version of BER called Generalized BER (GBER), which extends the original algorithm to sparse-reward environments, particularly those with complex structures that require the agent to explore. GBER improves the performance of BER by introducing relabeling mechanism and applying diverse sampling strategies. We evaluate our modified version, which is based on a goal-conditioned deep deterministic policy gradient offline learning algorithm, across various maze navigation environments. The experimental results indicate that the GBER algorithm can significantly boost the performance and stability of the baseline algorithm in various sparse-reward environments, especially those with highly structural symmetricity.

## I. INTRODUCTION

Reinforcement learning (RL) is a well-known potential solution for complex problems with less data compared to other supervised deep learning algorithms. From AlphaGo [1] to the latest update of the versatile ChatGPT-4 [2], from simple Atari games [3] to complex modern games [4], and from classic robotic arm motion control [5] to novel autonomous driving [6], RL has grown steadily, playing a critical role in deep machine learning.

Rewards are a critical component in the structure of RL [7]. Therefore, reducing the reward-shaping burden has been a popular topic in this field. Recently, a novel technique named back-stepping experience replay (BER), which proposes a simple but efficient mechanism to greatly enhance learning efficiency while reducing complex tasks of reward shaping, was published [8]. However, this study on BER was focused on a dense-reward environment, leaving a research gap between this novel technique and sparse-reward environments. A dense-reward environment refers to the environment where the agent can get non-zero reward signals in most situations. Reversely, in a sparse-reward environment, it is difficult for the agent to receive non-zero reward signals [7].

Thus, the reward problem is more crucial in sparse-reward environments. Without reward, nothing can be learned to optimize behavior, and failed exploration and sparse rewards can form a vicious cycle. Thus far, various methodologies have been proposed to overcome this issue. For example, Wu et al. [9] and Colas et al. [10] reported intrinsically motivated RL that uses curiosity or visual novelty as intrinsic rewards to reinforce the action of the agent; Andrychowicz et al. [11] proposed a goal relabeling strategy called hindsight experience replay (HER) to enable the agent to learn from failed experiences; Fang et al. [12], Luo et al. [13], and Nair et al. [5] employed algorithms that adopt more complex methodologies of decomposing difficult tasks into a series of simpler ones.

In this study, the original BER is modified into a new algorithm referred to as generalized back-stepping experience replay (GBER) to generalize it to sparse-reward environments. The major modifications focus on adopting relabeling mechanisms and diverse sampling strategies. The experimental results suggest that the GBER algorithm significantly boosts the performance of the baseline algorithm and stability in various sparse-reward environments.

The remainder of this paper is organized as follows: Section II reviews the related works. Section III presents some preliminaries of this study. Section IV introduces the modifications of GBER, and Sections V and VI present the experiments and analysis of experiment results, respectively. Section VI summarizes the conclusions of this study and discuss its limitations.

## II. RELATED STUDIES

This study is based on the framework of goal-conditioned reinforcement learning (GCRL), which is a subfield of RL wherein an agent is trained to pursue a single goal or a set of goal distributions [7, 14] and plans the goal itself to facilitate learning efficiency [15]. The first framework of GCRL can be traced back to the paper on universal value function approximators (UVFA) [16], which proposes a theory that an agent considers a goal (given or self-generated) and states to seek a policy for that goal.

The benefits of GCRL are as follows [14, 17]: (1) Designing reward functions is no longer challenging and is as simple as a binary function indicating whether an agent reaches the goal. (2) If the algorithm is well-designed, the agent can identify different goals with corresponding goal inputs, thereby making it possible to train a single agent for multiple tasks. (3) There are no strict rules for the forms of goals, which can be selected from among spatial coordinates [18], visual inputs [5], or even human language commands [19] as long as the algorithm can utilize the form.

However, the simplest binary reward function design still suffers from the sparse-reward problem. Given that states representing goal-related successes are significantly outnumbered by irrelevant states, all GCRL agents must navigate the exploration process in the absence of adequate rewards. To address this issue, HER utilizes failed experiences by relabeling the current achieved goals as successes [11]. HER enables an agent to learn from failures, similar to humans and other animals [11, 14]. Many other algorithms have been developed by expanding on HER. For example, algorithms to select useful experiences for training based on curriculum learning principles (e.g., proximity or difficulty of goals [12, 13, 20, 21]) or for choosing behavioral goals based on the achieved goals distribution (e.g., increasing the entropy of the distribution of all achieved goals [22]; replacing some goals in the buffer with the achieved goal closer to the other goals in the buffer [23]; using another neural network for generating new appropriate goals [20]).

The recent research on BER presents a new technique, viz., learning opposite skills by reversing existing experiences [8]. In BER, an agent is not only trained to reach a given task goal, say B, from its initial position, say A; however, it is also trained to reach A from B. In these two processes, the agent generates a track of virtual reversed experiences for each direction and learns from these virtual experiences and the real experiences simultaneously. The original BER was tested in a dense-reward environment. In contrast, this study focuses on sparse-reward environments that more extensive exploration.

## III. PRELIMINARIES
### A. GOAL-CONDITIONED REINFORCEMENT LEARNING

The framework of RL can be described as an interaction process between an agent and an environment. A typical environment comprises a state space $S$, an action space $A$, a reward function $r: S \times A \times S \to \mathbb{R}$, a transition probability $p(s_{t+1}|s_t, a_t)$, and a distribution of initial states $p(s_0)$. The entire process can be summarized as follows: The agent receives the observation of state $s_t$ and commits action $a_t = \pi(s_t)$, where $\pi$ indicates the decision-making policy of the agent. Subsequently, reward $r_t$ along with the next state $s_{t+1}$ is generated based on the reward function $r_t = r(s_t, a_t, s_{t+1})$ and transition probability $p(s_{t+1}|s_t, a_t)$. During the iterations of the previous processes, the agent gradually updates its policy $\pi$ by attempting to maximize the sum of future expected returns $R_t = \mathbb{E}[\sum_{i=t}^{T} \gamma^{(i-t)} r_i]$, where $T$ represents the terminal timestep and $\gamma$ represents the discount factor.

For GCRL, there is an additional factor, goal $g = \langle R_g, \gamma_g \rangle \in G$. $R_g: S \to \mathbb{R}$, $\gamma_g$, and $G$ represent a goal-conditioned reward function, corresponding discount factor, and goal space, respectively. In most cases, $S$ and $G$ are designed to be the same for convenience [14]. In GCRL, the agent performs action $a$ according to a goal-conditioned policy $\pi(s_t, g)$ [16]. $R_g$ takes the form of a binary function such as $\{1,0\}$ or $\{0,-1\}$ [24], which causes the GCRL to have a sparse-reward environment.

### B. HINDSIGHT EXPERIENCE REPLAY

In GCRL, the goal given by the environment at the beginning of each episode is referred to as a *desired goal*. The *achieved goal* is the state-corresponding goal achieved by the agent at the current state. In HER, some desired goals are replaced by the achieved goals in the same trajectories before updating the policy [14], which is referred to as *relabeling*. After relabeling, the agent is rewarded to reach the achieved goal successfully and uses this virtual success to optimize its policy. The most referred replay strategy in the original HER is "*future*," which refers to randomly picking an achieved goal in the future timesteps, and then committing relabeling and computing the new reward for 80% trajectories of the minibatch for updating [11]. The advantage of HER is that the agent can reuse many failed experiences and learn something from these experiences. However, it cannot fundamentally solve the sparse-reward problem of GCRL because the effect of HER relies on random explorations reaching task goals. If the task goals are too distant from the initial state, the likelihood of successful random explorations will decrease exponentially. Consequently, an agent employed with HER only learns from unsuccessful experiences.

### C. MAXIMUM ENTROPY GAIN EXPLORATION

In maximum entropy gain exploration (MEGA) [22], an agent attempts to solve the sparse-reward problem by setting special behavioral goals, which is a temporary replacement of overly difficult or far away desired goals. The agent picks an achieved goal from rarely visited areas in the distribution of previously achieved goals as a behavioral goal. MEGA can expand the distribution of the achievable area till it overlays the distribution of desired goals by employing this method.

The core algorithm of MEGA is deep deterministic policy gradient.

### D. "RFAAB"

In MEGA [22], the authors create a new diverse relabeling and sampling experience replay strategy named "rfaab" replacing the vanilla HER. The "rfaab" strategy refers to relabeling the current desired goal with a goal selected from five categories: *Real*, which refers to not relabeling at all; *Future*, which refers to the strategy of vanilla HER of the achieved goals in the future; *Actual*, which refers to a set of past desired goals; *Achieved*, which refers to a set of past achieved goals; and *Behavioral*, which refers to the current behavioral goal (no matter if it is the desired goal or not). "Rfaab" uses five numbers to express a set of hyperparameters representing the proportions for samples from each relabeling method. The most used hyperparameter "1_4_3_1_1" indicates that the proportion for "real," "future," "actual," "achieved," and "behavioral," respectively, is $1:4:3:1:1$. "Rfaab" is a strict generalization of the "future" strategy of vanilla HER. If transforming the vanilla HER "future" into the form of "rfaab," the hyperparameters would be "1_4_0_0_0." Through this strategy, MEGA demonstrates a nice performance in various environments such as FetchArm series, PointMaze2D, and AntMaze.

### E. BACK-STEPPING EXPERIENCE REPLAY

BER is inspired by the ability of humans to think both in the forward and backward manner [8]. BER imitates this mechanism by constructing a back-stepping transition for a standard transition, i.e., a reversed transition $(s_{t+1}, \widetilde{a}_t, s_t)$ for standard transition $(s_t, a_t, s_{t+1})$, in which $\widetilde{a}_t$ is calculated from an environment-dependent function $\widetilde{a}_t = f(s_t, a_t, s_{t+1})$. The reversed transition $(s_{t+1}, \widetilde{a}_t, s_t)$ should be similar to a real transition $(s_{t+1}, \widetilde{a}_t, s_{b,t})$, which means that $s_{b,t} \approx s_t$. Further, the desired goal is replaced by the corresponding goal of $s_0$ because of the change in direction.

The original BER contains two procedures: one is forward exploration from the initial position to the goal position, and the other is backward exploration from the goal position to the initial position. During each procedure, the standard and back-stepping trajectories are stored separately in two replay buffers. For optimization, a strategy $\mathbb{S}_t$ is used for collecting data from the two replay buffers. For the standard replay buffer $R_f$, the probability is $P_{t,f}$, and for the back-stepping replay buffer $R_b$, the probability is $P_{t,b}$, where $P_{t,f} + P_{t,b} = 1$. Subscript $t$ suggests that the probability $P_{t,b}$ gradually declines to zero for environments that are not perfectly reversible.

In addition, original BER addresses systems with partial reversibility by achieving approximate reversibility for choosing a simpler and solvable function $f$. However, this paper focuses on examining the mechanisms of BER in sparse-reward environments. This study chooses those environments that are perfectly reversible without the requirement of the reversibility approximator.

## IV. DISCUSSION AND MODIFICATION FOR BER
### A. ESSENCE OF BER

According to the original paper, BER is interpreted as "a bi-directional search method for standard off-policy RL approaches [8]." Further, in the same study, BER is applied to a model-free RL algorithm for evaluation and acquires some impressive results. We discuss its interpretation from the perspectives of relabeling strategy and human knowledge.

#### 1) RELABELING PERSPECTIVE

Relabeling is one of the major aspects of the operation of BER. HER relabels the current or future achieved goal as the desired goal, thereby computing a virtual reward with the generated data and learning from these failed experiences. BER relabels $s_t$ as the "next" state and $s_{t+1}$ as the "current" state. Meanwhile, it relabels $s_0$ as the desired goal and uses the same extrinsic reward function from the environment to calculate another virtual reward. Therefore, BER can be interpreted as a possible relabeling strategy similar to vanilla HER and "rfaab."

The two main differences between BER and HER are that BER relabels states and goals while HER only relabels goals, and BER requires generating new reversed actions $\widetilde{a}_t$ while HER does not. Therefore, BER can be considered a relabeling technique with an additional action data generation function. Further, BER and HER can be combined completely because of these differences.

#### 2) HUMAN KNOWLEDGE PERSPECTIVE

BER generates back-stepping transitions by sampling normal transitions and exchanges the position of $s_t$ and $s_{t+1}$. This process can be considered a type of data augmentation algorithm that generates new transitions based on human knowledge. However, there are several key points for distinguishing BER from typical Dyna-style method model-based RL data augmentation algorithms [25]. BER does not require training, and the role it plays during training is closer to reusing the environmental transition function based on human knowledge instead of learning a dynamics model from the environment [25]. Thus, BER can theoretically operate considerably faster than other typical learned dynamics models. In many environments, training several extra neural networks for environmental dynamics is unworthy, and executing them is another waste of time and energy. BER has an astonishingly low time complexity and energy cost as long as the user can find an appropriate function $f$. BER is not a learned limited environmental dynamics model; however, it utilizes an abstract rule summarized from human knowledge on the reversibility of environments. Owing to the previous two characteristics, the concept of BER can be expanded for constructing new algorithms to utilize various common rules from human knowledge on spacetime symmetry, such as spatial rotational reversibility.

### B. GENERALIZED BER

Despite numerous contributions, there are still some limitations in the original paper on BER:

(1) The original study is designed for specific dense-reward environments that do not require strong explorations.

(2) The backward exploration process is not necessarily efficient for more general environments and algorithms.

(3) The experience replay strategy can have more diversity via sampling more various relabeled transitions.

For (1), the original research designed BER for a soft snake robot in a simple goal-conditioned box environment. The reward function was complex and strongly related to the pairwise distances among the robot, goal, and initial position. The coefficients of the reward function required manual design. However, the back-stepping experiences were meaningful because of their dense-reward environments. We introduce HER into BER to examine the performance of BER in sparse-reward environments. This is one of the simplest methodologies for sparse-reward environments and suitable for comparing and examining the performance of the BER framework.

For (2), the original BER designs the backward exploration process for their specific environment; however, the target of this study is to discuss the core back-stepping mechanisms of BER in a more general scenario, wherein only a forward process is required. Further, we focus on environments with perfect reversibility, and we completely duplicate the backward exploration and back-stepping transitions. In addition, in the original BER, the agent relabels $s_0$ as the desired goal during the back-stepping transformation and uses a dense-reward function to train to reach it. For sparse-reward environments, these processes not only are very unlikely to generate useful rewards but also to reinforce the behavior around the initial state. Therefore, instead of relabeling the initial state $s_0$, adopting a reversed "future" strategy of HER may be a promising simple solution. The modified strategy must be revised to relabel the current desired goal with a randomly achieved previous goal that corresponds to states ranging from $s_0$ to $s_t$ because the previously achieved goals can be seen as the future achieved goals in reversed trajectories.

For point (3), this study implants other experience replay strategies. One crucial advantages of BER is the acquisition of opposite data without any substantial extra costs. Such data augmentation can significantly diversify the experience of the agent. However, there are many other proposals of sampling strategies to diversify the experience. Prioritized experience replay samples important transitions more frequently [26], curriculum-guided HER adaptively selects the failed experiences for replay partially based on diversity-based curiosity [12], and MEGA attempts to collect more experiences from those rarely visited areas [22]. Given the concise structure of BER, introducing another diversity-increasing experience replay mechanism will be the least destructive solution to its original framework. In this study, "rfaab" is introduced to BER as an example. BER can provide a virtual experience from the reverse direction to the agent, whereas "rfaab" can offer diverse existing experiences in the current direction. The two strategies constitute a perfect complement to each other. Besides "rfaab," there are many other possible relabeling and sampling methodologies that can be adopted by varying environments, which will be an appropriate aspect for future studies.

The entire framework of GBER can be summarized as the following pseudo-code.

**Algorithm 1** Generalized Back-stepping Experience Replay (GBER)

**Given:**
- an off-policy RL algorithm $\mathbb{A}$,
- a back-stepping action calculating function $f$,
- a reward function $r : \mathcal{S} \times \mathcal{A} \times \mathcal{G} \to \mathbb{R}$.
- a mapping function: $\mathcal{G} = m(\mathcal{S})$,
- a set of strategy $\{\mathbb{S}_1, \mathbb{S}_2, \mathbb{S}_3, ..., \mathbb{S}_n\}$ to sample goals for GBER relabeling,
- a set of sampling ratios $\{p_0, p_1, p_2, ..., p_n\}$ for the diverse relabeled replay buffers,

**Require:**
- approximate the reversibility of the system with function $f$

```
Initialize A                                    ▷ e.g. initialize neural networks
Initialize GBER replay buffer set {R_0, R_1, R_2, ..., R_n}
for episode = 1, M do
    Sample a goal g and an initial state s_0
    for t = 0, T − 1 do
        Sample an action a_t using the behavioral policy from A:   ▷ collecting transitions
            a_t ← π(s_t||g)
            r_t := r(s_t, a_t, g)
        Store the transition (s_t||g, a_t, r_t, s_{t+1}||g) in R_0
        Execute the action a_t and observe a new state s_{t+1}
    end
    for t = 0, T − 1 do
        Calculate the reversed action ā_t = f(s_t, a_t, s_{t+1})
        Sample a previously achieved goal for GBER replay ḡ := m[S_1(s_0, ...s_t)]
        r̄ := r(s_{t+1}, ā_t, ḡ)
        Store the transition (s_{t+1}||ḡ, ā_t, r̄, s_t||ḡ) in R_1
        Sample a set of goals for other relabel strategies:
            G_{2,3,4...n} := m[S_{2,3,4...n}(s_0, ...s_t, ...s_T)]
        Compute corresponding reward R_{2,3,4...n} := r(s_t, a_t, G_{2,3,4...n})
        Store the transitions (s_t||G_{2,3,4...n}, a_t, R_{2,3,4...n}, s_{t+1}||G_{2,3,4...n}) in R_{2,3,4...n}
    end
    for t = 1, N do
        Sample data on proportion {p_0, p_1, p_2, ..., p_n} from buffers {R_0, R_1, R_2, ..., R_n} and concatenate them into one minibatch B
        Optimize π using A and minibatch B
    end
end
```

## V. EXPERIMENTS
### A. ENVIRONMENTS

This study utilized most parts of MEGA as the basement algorithm and its original hyperparameters to conduct comparative studies among GBER, HER, and "rfaab" in various sparse-reward environments. Therefore, we use the *sibrivalry* maze environments tested in the baseline MEGA algorithm [22, 27].

The environments used in the study are AntMaze and PointMaze2D (Figure 1). **AntMaze** is a motion control environment, in which a four-leg robotic ant learns to control its limbs and joints to reach a goal through a U-shape maze. For the agent, the maze shape is unknown and invisible, and its observation is limited to its body status and task goal. **PointMaze2D** is a simple two-dimensional (2D) environment where a dot agent is required to move from its spawn position to the goal location. There is neither inertia nor friction in this environment. The action space is only on two axes. For the PointMaze2D series, BER is not only tested in the *"square_large"* maze, which is the main environment of the baseline MEGA algorithm [22], but also in two other mazes, namely, *Experiment_X_Y_Z* (X-unit wide and 2 × Y-unit long room with a thin wall at the Z-unit of Y-axis) and *Square_d* (a multi-goal maze with three branches).

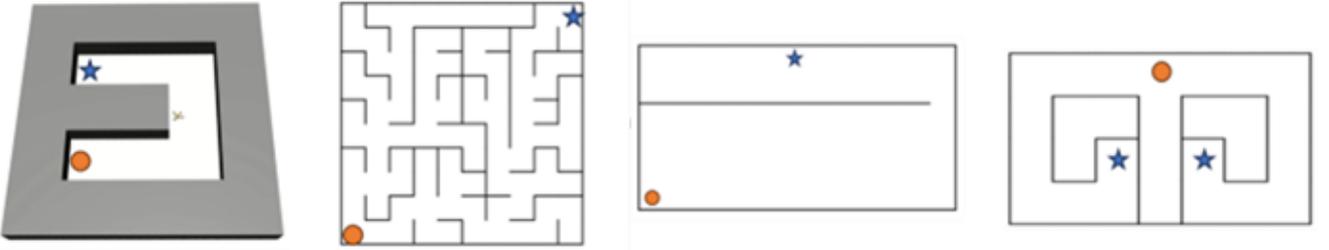

**FIGURE 1.** Environments used in the experiments. The orange circle is the spawn position of the agent, and the blue star is the task goal location. From left to right, the mazes are AntMaze, square_large, experiment_X_Y_Z, and square_d.

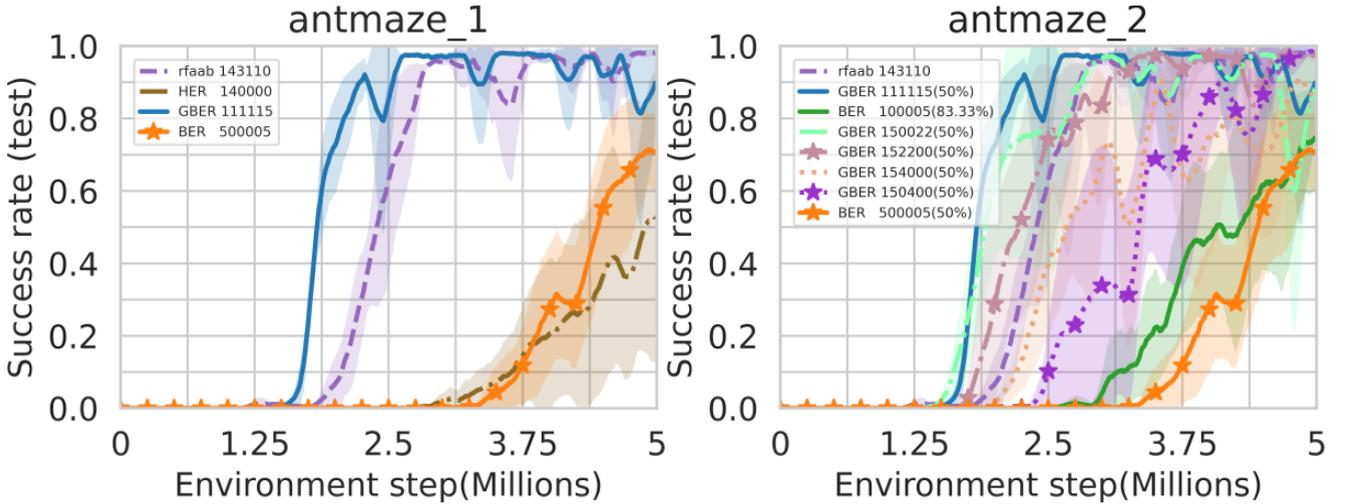

**FIGURE 2.** Separate plots of the Antmaze experiments. The percentages in the parentheses refer to the proportion of back-stepping transition samples.

### B. ALGORITHM SETTINGS

Considering the features of the experimental environments, the study selected some simple functions to test the potential of BER. For the back-stepping action generating function $f$, we selected $\tilde{a}_t = -a_t$. For comparison with the original BER without diverse sampling strategies and baseline algorithms without backward exploration processes, we removed the backward exploration part and decaying proportion for back-stepping transitions in the original BER paper. For the additional multiple experience replay module, this study employed "rfaab" as a benchmark for a comparative analysis. Each digit in the number after the name refers to the proportion of diverse relabeling strategies mentioned in Section III.D and the last digit refers to the proportion of back-stepping transitions. All results are fetched with the same baseline algorithm MEGA. The experiment is performed five times for each algorithm and the environment with five different seeds, which are shared for all algorithms. The success rate is evaluated every 5000 timesteps through ten tests.

### VI. RESULTS
### A. ANTMAZE

The results of Antmaze are presented in Fig. 2. The two plots are separated for clearer visualization. The first plot on the left indicates that GBER converges faster than pure "rfaab" (purple dash curve), which indicates that GBER can greatly enhance the overall performance of the baseline algorithm. In environments with perfect reversibility, the back-stepping transitions can be considered as "real" transitions with virtual rewards; however, it is to say that GBER can be seen as reversed HER. The results are that GBER and BER outperform HER with a considerably smaller proportion of relabeled transitions. The possible explanation is that the "U" shape of the maze requires the agent to learn two symmetric movement skills of routes A and C (Fig. 3). Owing to the fixed spawn position, the agent would have considerably more experiences of route A. However, GBER enables the agent to learn the skills for reversed route A (symmetric with route C) at the beginning and review the early skills via the same process in the later stage. Such an automatic balancing mechanism can be attributed to the trick of accelerating learning, which can explain why the GBER curve converges only a little earlier than "rfaab," and has a steeper slope. The GBER agent has already learned something opposite at the beginning, and it can reach the goal with this knowledge.

The second plot in Figure 2 shows that the major finding is the importance of diversity in relabeled transition samples. The curve "GBER 111115," which has the most diverse samples of various relabeling strategies, beats all other patterns, indicating the highest converging speed and average performance level. Indeed, these results concretely show that the diversity of relabeled samples is one of the most important reasons affecting the performance of GBER.

In summary, GBER shows a strong potential for improving the baseline algorithm MEGA in AntMaze. It helps the baseline algorithm converge faster and reach higher levels with almost negligible extra costs.

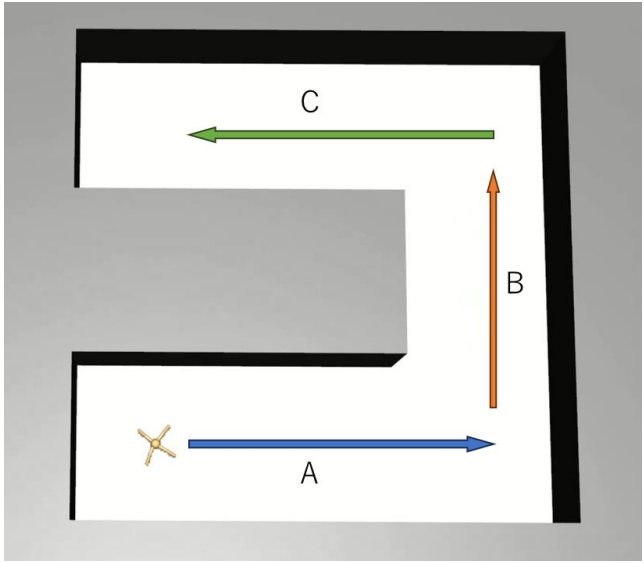

**FIGURE 3.** Schematic of the movement route of the Ant agent.

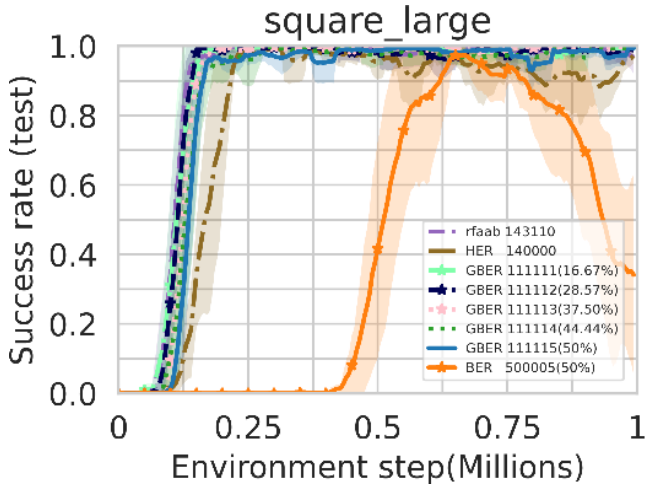

**FIGURE 4.** Results of "square_large" in PointMaze2D environments. The shadow area indicates the range of results.

### B. PointMaze2D

#### 1) SQUARE_LARGE

Figure 4 shows that the original BER has the worst result and it immediately faces a catastrophic decline after it reaches its highest success level. Subsequently, various GBER and "rfaab" curves have the best results. The BER has a considerably larger range of results (shadow area) compared to the results of GBER and "rfaab." The possible interpretation is that diversity not only increases the converging speed of the original BER, but also increases its stability.

#### 2) EXPERIMENT_X_Y_Z

In Fig. 5, both GBER and "rfaab" have similar performances at the first 250,000 timesteps.

However, the "rfaab" faces a catastrophic decline. Simultaneously, GBER presents a stable performance as before. The orange line of BER has a similar slope to the others; however, its performance is considerably worse, and the range of its results is also large. Further, the range of the "rfaab" results increases after the decline begins, whereas GBER has a considerably smaller range of performance in all periods. The possible explanation is that GBER combines diverse virtual experiences with generated back-stepping transitions from the other direction to exploit both, resulting in a considerably better and stabler performance.

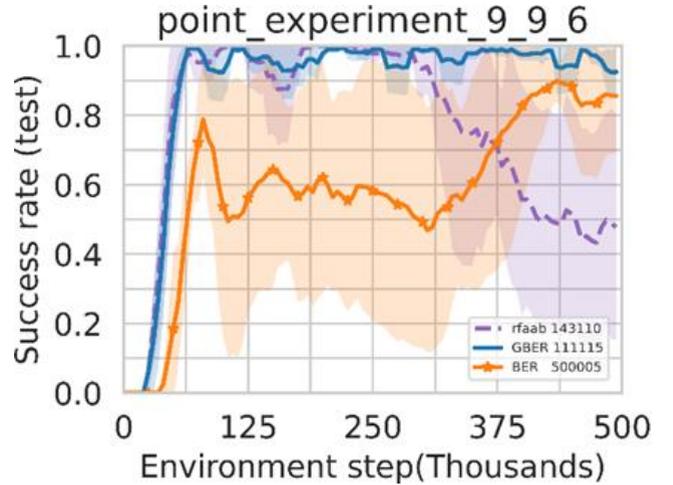

**FIGURE 5.** Results of "experiment_9_9_6" in PointMaze2D environments. The shadow area indicates the range of results.

#### 3) "SQUARE_D"

"Square_d" is an axisymmetric maze with the spawn position in the middle and two task goals on both sides. As expected, additional GBER samples help the baseline algorithm MEGA learn faster and better than both "rfaab" and BER (Figure 6). The can be attributed to the highly structural symmetricity of the environment, which is similar to the AntMaze case.

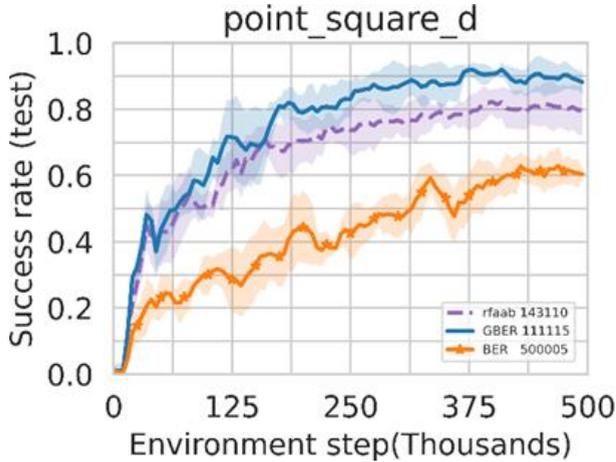

**FIGURE 6.** Results of "Square_d" in PointMaze2D environments.

## VII. CONCLUSIONS AND DISCUSSIONS

Herein we proposed an improved version of BER with high modifiability, viz., GBER. The improved version of BER distinguishes itself from the original version via generalization to sparse-reward GCRL environments, adoption of relabeling mechanisms, and experience replay sampling diversity. With the simplest operations, GBER introduces back-stepping transitions to the baseline algorithm and improves its performance in most environments, especially those with some structural symmetricity. Further, by comparing the results of the different proportions of relabeling strategies, we showed that the sampling diversity of GBER helps outperform the agent employed with the original BER. Moreover, GBER enhanced the stability of performances and avoided the possible catastrophic decline.

Some limitations need to be addressed for future studies:

(1) The current GBER could not improve the performance of the baseline algorithm in over-asymmetric and over-complex environments.

(2) This study examined perfectly reversible sparse-reward environments. The best proportion of GBER and other strategies can vary based on different environments, which requires many tests and manual adjustments.

(3) There may be many other strategies that can be incorporated with GBER. Further, with an increase in new strategies, the second and the third problem is expected to become more complicated.